\documentclass[runningheads]{llncs}
\usepackage[T1]{fontenc}
\usepackage{amssymb,multirow}
\usepackage{graphicx,verbatim,amsmath}
\usepackage{float}
\usepackage{xcolor}
\usepackage{soul}
\begin{document}
\title{F2PASeg: Feature Fusion for Pituitary Anatomy Segmentation in Endoscopic Surgery}
%

\author{Lumin Chen\textsuperscript{1},
Zhiying Wu\textsuperscript{1*}, 
Tianye Lei\textsuperscript{2}, 
Xuexue Bai\textsuperscript{3}, 
Ming Feng\textsuperscript{3}, 
Yuxi Wang\textsuperscript{1}, 
Gaofeng Meng\textsuperscript{1}, 
Zhen Lei\textsuperscript{1}, 
and Hongbin Liu\textsuperscript{1}}  
\authorrunning{L. Chen et al.}
\institute{
    \textsuperscript{1}Centre for Artificial Intelligence and Robotics, Hong Kong Institute of Science \& Innovation, Chinese Academy of Sciences \\
    \email{zhiying.wu@cair-cas.org.hk} \\
    \textsuperscript{2}The University of Hong Kong \\
    \textsuperscript{3}Peking Union Medical College Hospital, Chinese Academy of Medical Science     
}

\maketitle              
\begin{abstract}
Pituitary tumors often cause deformation or encapsulation of adjacent vital structures. Anatomical structure segmentation can provide surgeons with early warnings of regions that pose surgical risks, thereby enhancing the safety of pituitary surgery. However, pixel-level annotated video stream datasets for pituitary surgeries are extremely rare. To address this challenge, we introduce a new dataset for \underline{P}ituitary \underline{A}natomy \underline{S}egmentation (PAS). PAS comprises 7,845 time-coherent images extracted from 120 videos. To mitigate class imbalance, we apply data augmentation techniques that simulate the presence of surgical instruments in the training data. One major challenge in pituitary anatomy segmentation is the inconsistency in feature representation due to occlusions, camera motion, and surgical bleeding. By incorporating a \underline{F}eature \underline{F}usion module, F2PASeg is proposed to refine anatomical structure segmentation by leveraging both high-resolution image features and deep semantic embeddings, enhancing robustness against intraoperative variations. Experimental results demonstrate that F2PASeg consistently segments critical anatomical structures in real time, providing a reliable solution for intraoperative pituitary surgery planning. Code: https://github.com/paulili08/F2PASeg.


\end{abstract}
\section{Introduction}
Automatic segmentation of anatomical structures can identify dangerous areas, surgical risks can be reduced in pituitary surgery~\cite{patel2016skull,van2020endoscopic,wang2015endoscopic,DBLP:conf/miccai/XuWCCLL24,DBLP:journals/corr/abs-2311-09965,DBLP:conf/iros/ChenZGL0W0L24}. Especially during the sella phase, anatomical structure segmentation is crucial due to the close proximity of the anatomy~\cite{patel2016skull}. As dangerous areas are difficult to segment, surgeons face challenges when performing endoscopic pituitary surgery. The sella, where the pituitary tumor is located, can be accessed safely. However, the presence of internal carotid arteries and optic nerves beneath the smaller surrounding structures complicates the process of opening. The pituitary tumor leads to compression, distortion, or encasement of the surrounding structures~\cite{bonneville2016magnetic}. Inaccurate segmentation of essential anatomical structures can cause injury to the patients~\cite{marcus2021pituitary,patel2016skull,FT-Reg2025,MRtoCT2025isbi}.

\begin{figure*}[t]
  \centering
  \includegraphics[width = \textwidth]{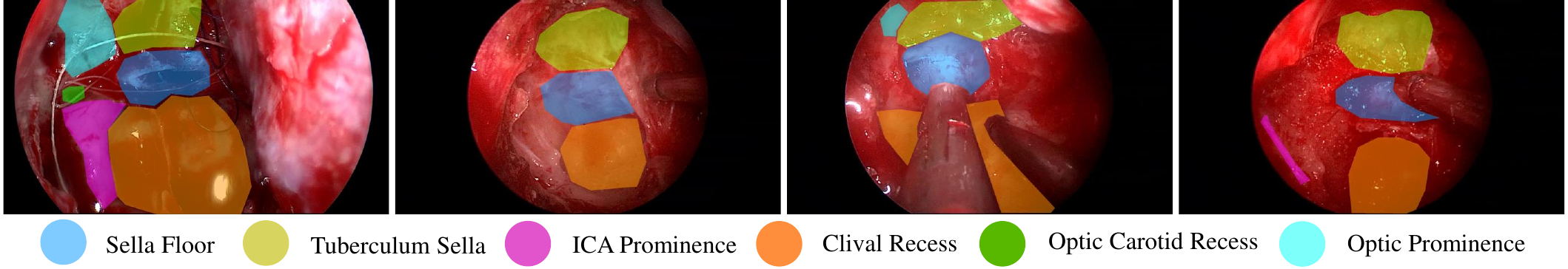}   \\
  \caption{Examples of the proposed PAS dataset including images and labels.}
  \label{fig1}
\end{figure*}

The availability of anatomical datasets for pituitary surgeries is extremely limited. The scarcity of comprehensive and diverse datasets in pituitary surgeries poses a significant challenge during the sellar phase. Due to the considerable variations in anatomical structures among patients, collecting a large-scale dataset is crucial. Fig.~\ref{fig1} illustrates the semantic segmentation of six anatomical structures during the sellar phase of endoscopic pituitary surgery. As shown in Table~\ref{tab:data_sample}, our dataset consists of 7,845 images. Compared to~\cite{sarwin2023live} and~\cite{staartjes2021machine}, bounding boxes/centroids annotated datasets produced for target detection task, our dataset provides pixel-level mask annotations for semantic segmentation, which is much more labor-intensive. Specifically, in the sellar phase, the number of images in our dataset is 12 times greater than that of the dataset in~\cite{das2023multi}. Moreover, our dataset includes nearly twice as many cases as the dataset presented in~\cite{das2023multi}, with images in each case exhibiting a high degree of continuity, making them suitable for video-based analysis. Additionally, our dataset captures the significant variations in anatomical structures among patients and provides a comprehensive representation of the diverse anatomical scenarios encountered in pituitary surgeries.

\begin{table}[t]
\centering
\caption{Comparison with the existing datasets in pituitary anatomy segmentation.}
\label{tab:data_sample}
\centering
\renewcommand{\arraystretch}{1}
\setlength{\tabcolsep}{6pt}
\begin{tabular}{c|c|c|c|c}
\hline
Pituitary Anatomy Datasets & Image & Case & Class & Task\\
\hline
Sarwin et al.~\cite{sarwin2023live} & 19000 & 166 & 16 & Detection \\ 
Staartjes et al.~\cite{staartjes2021machine} & 549 & 23 & 3 & Detection \\\hline
Adrito et al.~\cite{das2023multi} & 635 & 64 & 10 & Segmentaion\\
Our PAS &  7845 & 120 & 6 & Segmentaion\\
\hline
\end{tabular}
\end{table}

A number of commonly used deep learning methods have been used for intraoperative endoscopic segmentation~\cite{hao2020brief,wang2015endoscopic,wu2024surgivisor}. U-Net~\cite{ronneberger2015u} uses weakly supervised learning of centroids to generate segmentation masks for each structure. In~\cite{das2023multi}, U-Net++ is employed for the semantic segmentation of the two most prominent, largest, and frequently occurring structures (sella and clival recess) and for centroid detection of the remaining eight less frequently occurring structures. Recently, the field of semantic segmentation has undergone a significant shift with the increasing focus on large-scale pre-trained models. Segment Anything Model (SAM)~\cite{kirillov2023segment}, a leading Vision Transformer-based segmentation framework, has made remarkable progress in expanding the boundaries of segmentation in natural images. The subsequent update of Segment Anything Model 2 (SAM2)~\cite{ravi2024sam} enables efficient video segmentation by transferring prompts with frame-to-frame continuity. It has largely achieved the end-to-end efficient segmentation required for intraoperative endoscopy, the current segmentation methods have not fully investigated feature fusion. Thus, we purpose F2PASeg, a video-based segmentation model enhancing feature fusion for complex scenes. The main contributions are summarized as follows:

\begin{enumerate}
    \item We introduce the \underline{P}ituitary \underline{A}natomy \underline{S}egmentation (PAS) dataset, a large-scale collection consisting of 7,845 pixel-level annotated images captured during the sellar phase of endoscopic pituitary surgery.  Our dataset PAS contains the significant variations in anatomical structures among patients.
    \item We propose an efficient architecture F2PASeg for pituitary anatomy segmentation in endoscopic surgery. In our F2PASeg, a feature fusion module enhances the mask decoder's abilities to integrate image embeddings with high-dimensional features from the image encoder to optimize feature integration.
    \item To address imbalanced distributions of critical structures, we multiplex surgical instrument annotations from the same dataset for data augmentation. In particular, for the sparsely distributed categories of pituitary anatomy, the original image frames are augmented with simulated surgical scenes that involve the use of surgical instruments.
\end{enumerate}

\section{Methods}
\subsection{Overview}
In this section, we build an end-to-end promptable structure F2PASeg and augment the dataset for pituitary tumor surgery scenarios. First, we integrate a feature fusion module in mask decoder. This module combines two residual blocks with LoRa branch, which enhances the combinations between the features from image encoder and the embedding from memory encoder. With LoRA, the model better satisfies intraoperative real-time segmentation demands, achieving higher FPS and reduced parameters. Second, we augment for samples with small distributions by multiplexing the dataset to target less distributed samples, mitigating the effect of imbalanced distributions.

\begin{figure*}[t]
  \centering
  \includegraphics[width =0.9\textwidth]{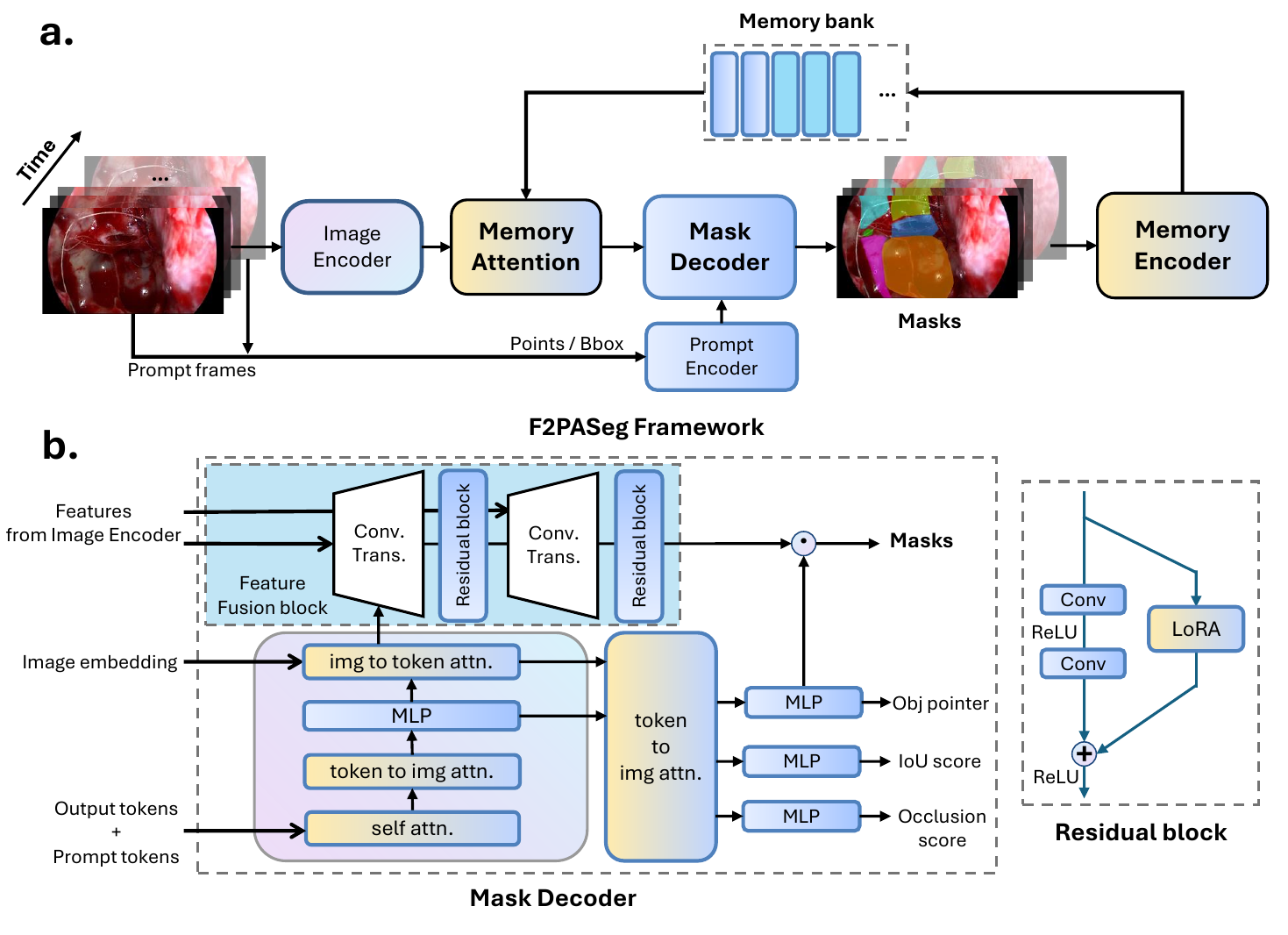}   \\
  \caption{The structure of our proposed F2PASeg model. (a) is the overall framework that contains image encoder, prompt encoder, memory attention, mask decoder and memory encoder. (b) is the modified mask decoder containing a feature fusion module with two residual blocks in parallel with attention branch. The residual block has an additional LoRA branch.}
  \label{framework}
\end{figure*}

\subsection{F2PASeg backbone}
The Segment Anything Model (SAM)~\cite{kirillov2023segment} has demonstrated strong performance in image segmentation tasks. However, its heavy reliance on prompts makes it unsuitable for intraoperative endoscopic surgery scenarios. The later version, SAM2~\cite{ravi2024sam}, extends image segmentation to the video domain and generates mask predictions across an entire video by leveraging a newly developed memory mechanism. 
As illustrated in Fig.~\ref{framework}(a), the proposed F2PASeg incorporates three memory-centric architectural innovations: a memory encoder, memory bank, and memory attention module. Specially, the memory bank implements a FIFO queue system that stores both recently predicted frames and prompt frames, capturing spatial feature maps and object pointers to maintain temporal semantic information. To prevent cross-scene interference, we implement a prompt partitioning mechanism that selectively stores only the two most recent prompt frames alongside non-prompt predicted frames in the memory bank. The memory mechanism enables the model to effectively incorporate historical predictions and supplementary prompts into the current frame's feature processing.

Similarly, the mask decoder is modified to align with the new memory mechanism. The output of memory attention integrates high-resolution information from the hierarchical image encoder using two transposed convolution blocks as skip connections. However, the original SAM2 network directly adds high-dimensional features to embeddings, which does not effectively leverage this meta information. Prior studies~\cite{chobola2023miccai,zhao2017deep} have shown that incorporating residual blocks in the feature fusion process improves feature integration. Therefore, we introduce residual computation when fusing features at strides 4 and 8 to enhance feature refinement.

Specifically, we take the high-resolution features $\mathbf{F}_{\text{high}}$   and the output from memory attention after transposed convolution $\mathbf{F}_{\text{mem}}$ as inputs to a residual block (Fig.~\ref{framework}(b)). The residual fusion process is formulated as:

\begin{equation} \mathbf{F}_{\text{res}} = \sigma \left( \mathcal{H}(\mathbf{F}_{\text{high}}) + \mathbf{F}_{\text{mem}} \right) \end{equation}
where $\mathcal{H}(\cdot)$ represents a sequence of operations applied to $\mathbf{F}_{\text{high}}$, including convolution, batch normalization, and ReLU activation $\sigma(\cdot)$. Besides, we add a Low-Rank Adaptation (LoRA) branch parallel to the main convolutional path:

\begin{equation} 
\mathbf{F}_{\text{res}}^{'} = f\left( \mathbf{F}_{\text{res}}\right) + \alpha \mathbf{B} (\mathbf{A} \mathbf{F}_{\text{res}}) 
\end{equation}
where \(\mathbf{A} \in \mathbf{R}^{r \times d}\) and \(\mathbf{B} \in \mathbb{R}^{d \times r}\) are the low-rank projection matrices in the LoRA branch. $r$ is the rank of the decomposition, typically much smaller than $d$ to keep computations efficient. $\alpha$ is a scaling factor that controls the strength of the LoRA adaptation. The LoRA term \(\mathbf{B}(\mathbf{A} \mathbf{F}_{\text{res}})\) provides an additional feature modulation pathway, allowing feature modulation without modifying the entire model, making it lightweight and flexible and leading to better spatial-temporal fusion.

The proposed model combines weighted focal loss, dice loss, mean-absolute-error (MAE) loss and cross-entropy (CE) loss. The over all loss function is given as follows:
\begin{equation}
    Loss = \lambda_{focal}Loss_{focal} + \lambda_{Dice}Loss_{Dice} + \lambda_{MAE}Loss_{MAE} + \lambda_{CE}Loss_{CE}
\end{equation}
where $\lambda_{focal}$, $\lambda_{Dice}$, $\lambda_{MAE}$, and $\lambda_{CE}$ are the weights of each loss, respectively. The values are set to $20:1:1:1$, respectively, according to~\cite{ravi2024sam}.

\begin{figure}[bt]
    \centering
    \includegraphics[width=\linewidth]{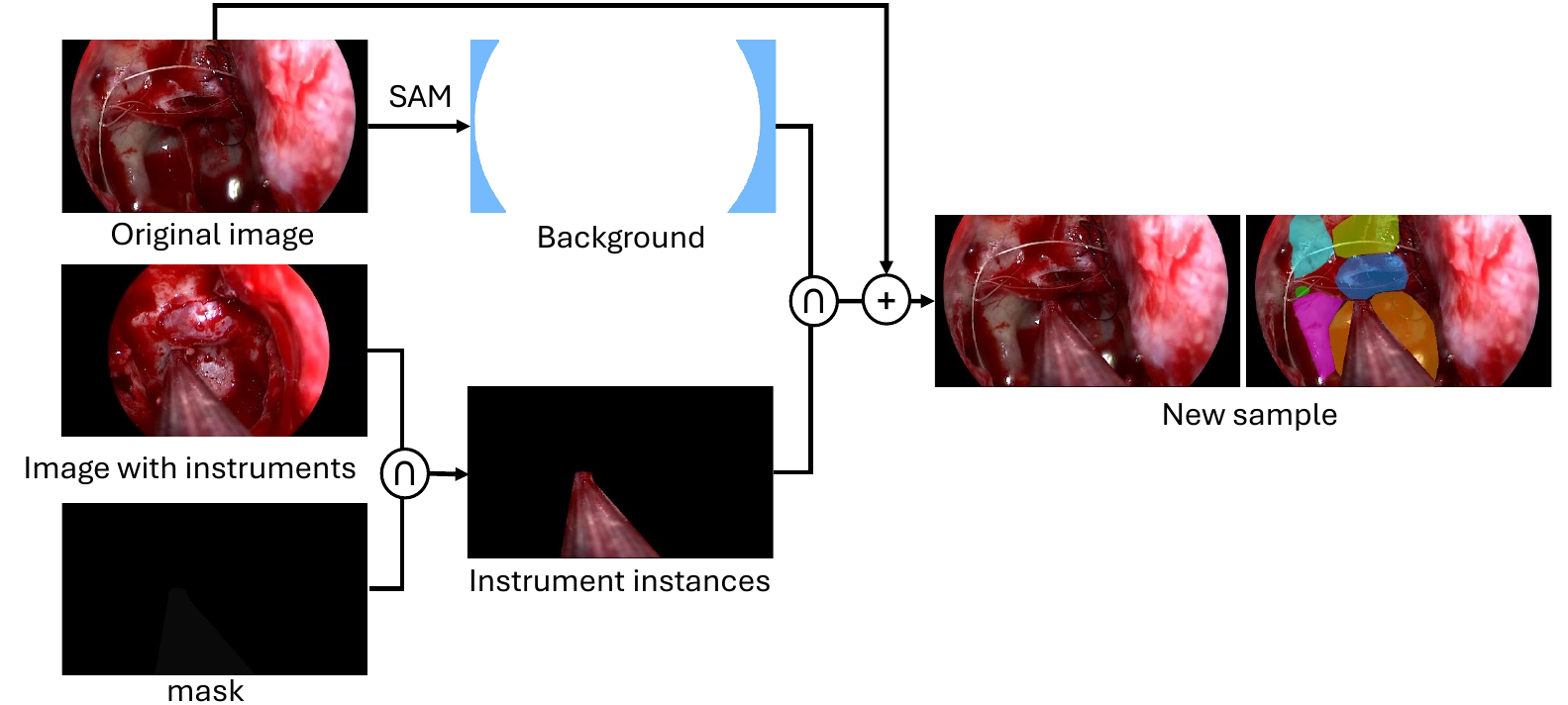}
    \caption{Data augmentation pipeline.}
    \label{fig:data_aug}
\end{figure}

\subsection{Data Augmentation}
In our dataset, there are three larger and more distinct structures presenting in all videos: sella floor (SF), tuberculum sella (TS) and clival recess (CR). However, ICA prominence (IP), optic carotid recess (OCR), and optic prominence (OP) are significantly scarcer. Despite their low prevalence, these structures are crucial in pituitary tumor surgeries, as injuries to the internal carotid artery (ICA) and optic nerve can result in hemorrhage or vision impairment~\cite{arnold2022pituitary,madani2022artificial}.
As shown in Fig.~\ref{fig:data_aug}, we implement a video reuse approach for annotating eight surgical instruments that are frequently employed during the sellar phase, as determined by expert neurosurgeons. The annotated instruments comprise: suction tube, rongeur, cutting forceps, cup forceps, bipolar electrode, freer, and scissors. We then select the cases containing ICA and OCR and superimpose surgical instruments from the additional cases into the original images in chronological order. This augmentation simulates realistic surgical scenarios, incorporating occlusions and motion artifacts caused by instruments.

\begin{figure}[t]
    \centering
    \includegraphics[width=0.75\linewidth]{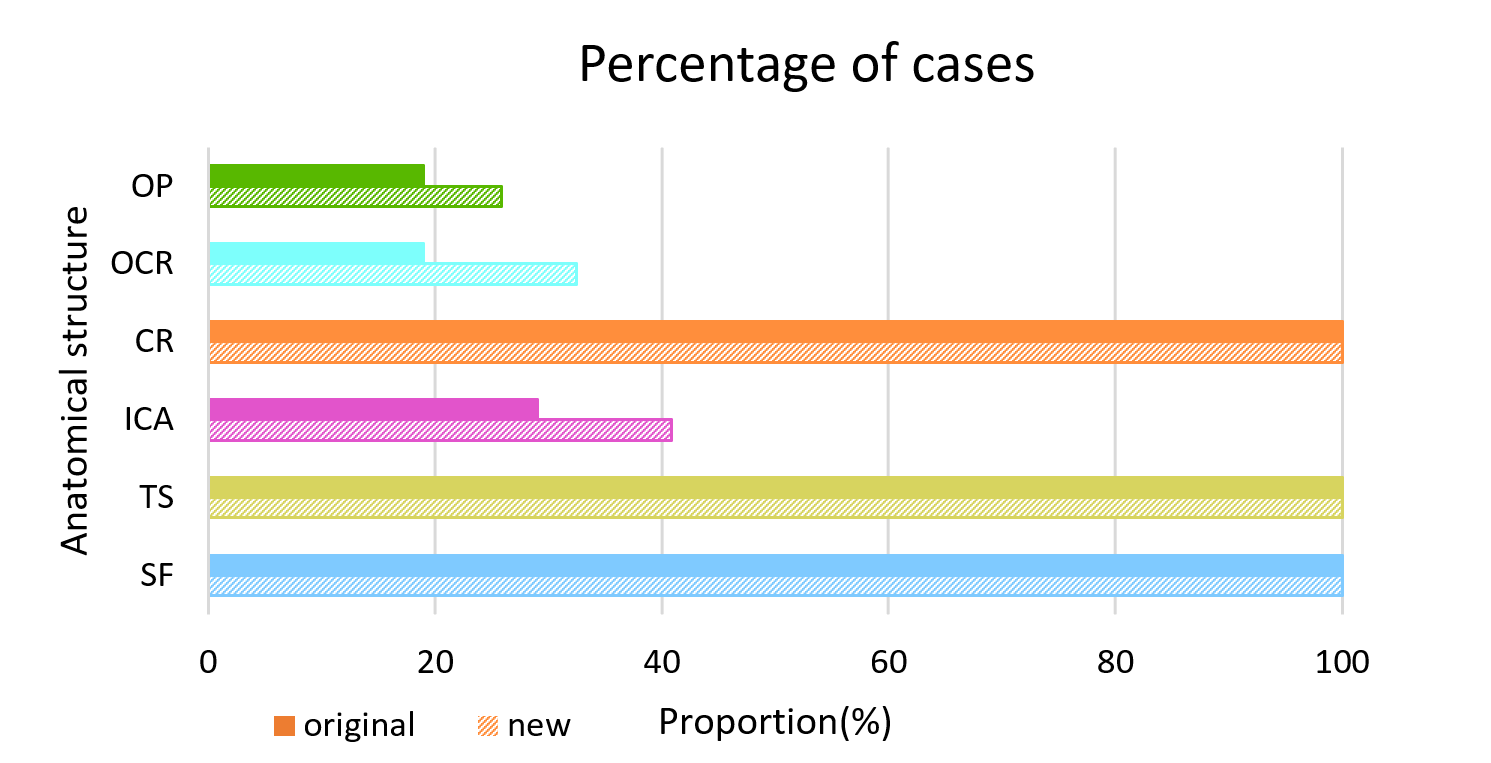}
    \caption{Distribution of 6 anatomical structures in dataset.}
    \label{fig:data_proportion}
\end{figure}

\section{Experiment}
\subsection{Dataset Description}
Our dataset comprises 7,845 images extracted from 120 videos of endoscopic pituitary surgery during the sellar phase, with each frame having a resolution of either $1920\times1080$ or $720\times576$ pixels. The anatomical structures in each frame are categorized into six classes: sella floor (SF), tuberculum sella (TS), ICA prominence (IP), clival recess (CR), optic carotid recess (OCR), and optic prominence (OP). All images are annotated by specialized neurosurgeons, with a small subset labeled by researchers and later reviewed by neurosurgeons for accuracy.
We first choose 100 cases and split them into $70$ cases for training, $10$ cases for validation, and $20$ cases for testing, ensuring a balanced distribution for model evaluation. Compared to previous works~\cite{das2023multi,staartjes2021machine}, our dataset offers more training images and higher resolution, providing a more comprehensive representation of anatomical variations. Fig.~\ref{fig:data_proportion} illustrates the proportions of the six anatomical structures across the 100 cases in the dataset. After data augmentation, the proportions of key structures increase significantly, with ICA reaching $40.83\%$ and OCR $32.50\%$, greatly alleviating the imbalanced distribution of samples. As a result, the dataset expands to 9,331 images, with 88 cases for training and 12 for validation, and 20 cases remain unchanged for testing.

\subsection{Implementation Details}
We fine-tune our model based on SAM2-t pretrained weight and setting and reduce the number of prompt frames in the memory bank. The mask decoder is frozen, while all other components remain trainable. Bounding box prompts are provided for each anatomical structure every 10 frames. Input images are processed at multiple scales, with the image encoder generating 1024-resolution outputs. The implementation is based on Python 3.12.8 and PyTorch 2.5.1, running on two NVIDIA A100 Tensor Core GPUs with CUDA 12.4. Training is conducted for 40 epochs, and the best model is obtained using the AdamW optimizer($\beta_1=0.9$, $\beta_2=0.999$) with a base learning rate of $5.0 \times 10^{-6}$.

\begin{table*}[t]
\centering
\caption{Quantitative Comparison of different models}\label{tabresult}
\resizebox{0.75\textwidth}{!}{
\begin{tabular}{c|c|c|c|c|c|c|c|c}
\hline
  \multirow{2}{*}{Model} & \multirow{2}{*}{mIoU} &  \multicolumn{7}{c}{Dice} \\
  \cline{3-9}
  & & Mean & SF & TS & IP & CR & OCR  & OP \\
\hline

 Swin-UNet~\cite{cao2022swin} &{0.1872}&{0.2509}&{0.6360}&{0.3650}&{0.0114}&{0.4590}&{0.0121}&{0.0216}\\
 Trans-UNet~\cite{chen2021transunet} &{0.2192}&{0.2847}&{0.7247}&{0.4322}&{0.0222}&{0.4806}&{0.0002}&{0.0483}\\
 DeepLabV3+~\cite{chen2018deeplabv3+} &{0.2085}&{0.2434}&{0.7500}&{0.0511}&{0.0002}&{0.5599}&{0.0017}&{0.0958}\\
 LiVOS~\cite{liu2025livos} &{0.4264} & {0.5057} & {0.8210} & {0.5851} & {0.2627} & {0.6752} & {0.1609} & {0.5293}\\
 SAM~\cite{kirillov2023segment} &{0.6090} & {0.7188} & {0.8280} & {0.6468} & {0.5988} & {0.7169} & {0.7252} & {0.7970}\\
 SAM-Med 2D~\cite{cheng2023sam} &{0.6648}&{0.7798} &{0.8600}&{0.8403}&{0.6087}&{0.8076}&{0.7106}&{0.8514}\\
 MedSAM~\cite{ma2024segment} &{0.7086} & {0.8166} & {0.8641} & {0.8407} & {0.7368} & {0.8119} & {0.7792} & {0.8670}\\
 SAM2~\cite{ravi2024sam} &{0.7681}&{0.8397}&{0.9043}&{0.8757}&{0.7301}&{0.8667}&{0.8049}&{0.8564}\\
 Ours &{0.7701}&{0.8559}&{\textbf{0.9158}}&{\textbf{0.8917}}&{0.7431}&{0.8826} &{0.8133}&{0.8887}\\
 Ours (+Aug) &{\textbf{0.7796}}&{\textbf{0.8635}}&{\textbf{0.9158}}&{0.8901}&{\textbf{0.7821}}&{\textbf{0.8860}}&{\textbf{0.8181}}&{\textbf{0.8888}}\\
\hline
\end{tabular}
}
\end{table*}

\subsection{Results}
\begin{figure*}[t]
  \centering
  \includegraphics[width = 0.95\textwidth]{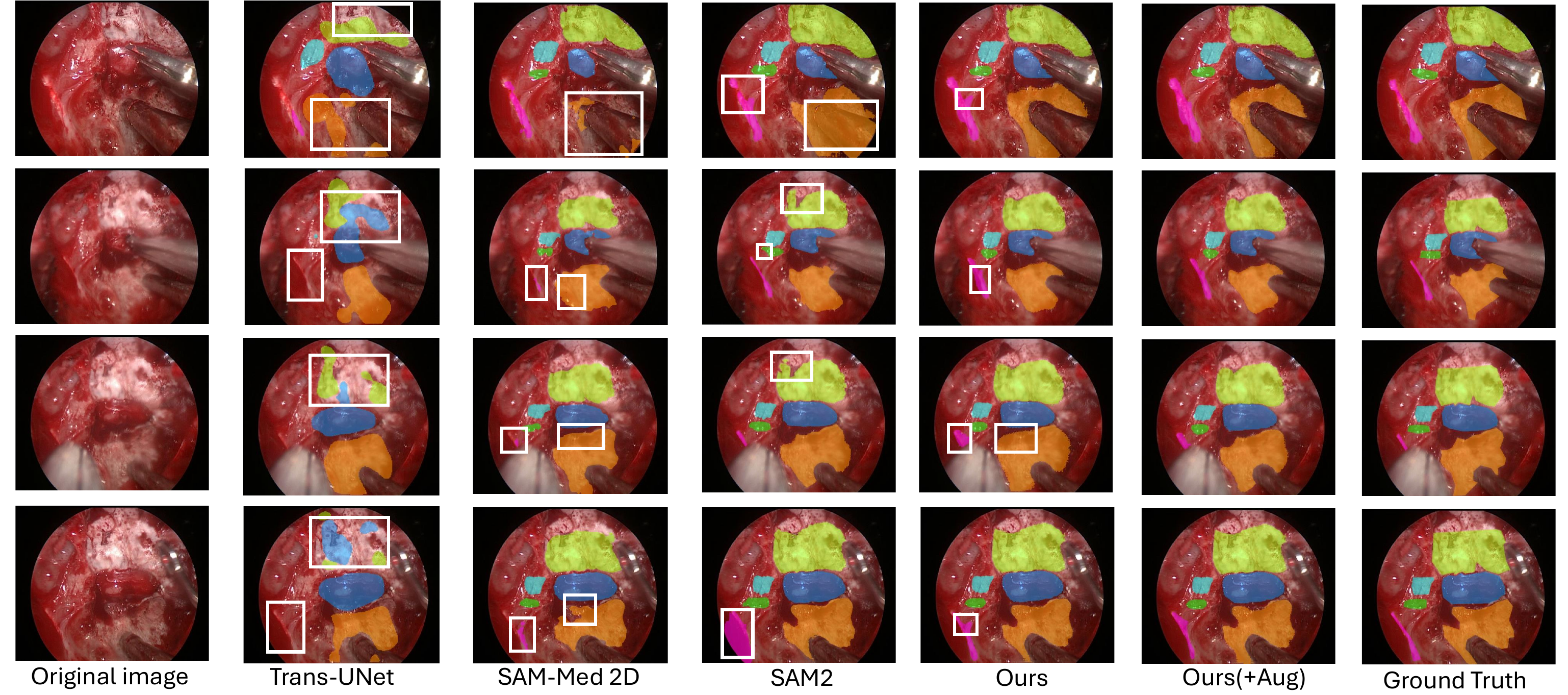}   \\
  \caption{visualization result}
  \label{result}
\end{figure*}

To evaluate the effectiveness of our model, we train unprompted image models Swin-UNet~\cite{cao2022swin}, Trans-UNet~\cite{chen2021transunet},  DeepLabV3+~\cite{chen2018deeplabv3+}, video segmentation model LiVOS~\cite{liu2025livos}, fully-prompted model SAM-Med 2D~\cite{cheng2023sam}, MedSAM~\cite{ma2024segment} and the original SAM~\cite{kirillov2023segment} and SAM2~\cite{ravi2024sam} and compare their performance. As shown in Table~\ref{tabresult}, our F2PASeg achieves a mean Dice score of 85.59\%, which is 1.62\% higher than the original SAM2 and 4.69\% higher than the fully-prompted MedSAM. Due to the complexity of the scene, the unprompted models perform poorly and can only detect three more obvious structures. Fig.~\ref{result} provides a visual comparison of segmentation results. F2PASeg demonstrates superior segmentation in dynamically changing regions, particularly those affected by bleeding or instrument occlusion during surgery. In particular, in the second and fourth columns of Table~\ref{tabresult}, TS (top of frame) and CR (bottom of frame) are often affected by camera pans. The feature fusion module improves segmentation continuity, ensuring that predictions remain more stable and aligned with the prompt frames. Moreover, F2PASeg achieves 28.57 FPS, which is 2.3 times higher than that of SAM-Med 2D, and basically meets the requirement of intraoperative real-time segmentation. With the LoRA module, the number of training parameters decreased from 39.0M to 34.8M.

\subsubsection{Ablation Studies}
To verify the validity of our designed model and data augmentation strategy, we conduct the ablabtion studies. The detailed results with different configurations are listed in Table~\ref{tabAb}. These results indicate that our model enhances segmentation performance by effectively modeling the relationships between anatomical structures within the feature fusion module. Additionally, the third row of Table~\ref{tabresult} presents detailed results, revealing segmentation accuracy improvements of 3.90\% for ICA and 0.48\% for OCR compared to the previous model. Notably, F2PASeg reduces incorrect segmentation of surgical instruments relative to the original model. Additionally, our data-augmented model further enhances the segmentation accuracy of ICA, reinforcing the efficiency of our spatial feature extraction method.

\begin{table}[t]
\centering
\caption{Ablation Studies Results}\label{tabAb}
\centering
\resizebox{0.6\textwidth}{!}{
\begin{tabular}{c|c|c|c}
\hline
  Feature Fusion Module & Data Augmentation & mDice & mIoU \\
\hline
- & - & {0.8397} & {0.7681}\\
 - & \checkmark &{0.8531} & {0.7697}\\
 \checkmark & - &{0.8559} & {0.7701}\\
\checkmark & \checkmark &{\textbf{0.8635}} & {\textbf{0.7796}}\\
\hline
\end{tabular}
}
\end{table}

\section{Conclusion}
In this paper, we addressed pituitary anatomy segmentation during the sellar phase of pituitary surgeries. We introduced a large-scale dataset, Pituitary Anatomy Segmentation (PAS), comprising 7,845 high-resolution, temporally coherent images from 120 surgeries. Each image has been meticulously annotated by expert neurosurgeons. We proposed F2PASeg, an efficient architecture designed to explicitly model relationships between anatomical structures in endoscopic surgery. Our method achieved a mean Dice score of 86.35\%. The segmentation accuracy for carotid arteries, a critical structure for surgical safety, was notably enhanced. This improvement provides early warnings of high-risk regions, assisting surgeons in their procedures. In addition, F2PASeg meets the real-time processing requirements for video-based intraoperative applications. This ensures seamless integration into surgical workflows, providing live, high-accuracy anatomical segmentation during endoscopic procedures. As future work, we plan to explore adaptive temporal modeling to further enhance segmentation robustness in long video sequences.

\begin{credits}
\subsubsection{\ackname} 
The research in this paper was funded by Inno HK program.
\subsubsection{\discintname}
The authors have no competing interests to declare relevant to this article’s content.
\end{credits}
%
%
%
%
\bibliographystyle{splncs04}
\bibliography{Paper-1527}
\end{document}